\documentclass[10pt,twocolumn,letterpaper]{article}

\usepackage{cvpr}
\usepackage{times}
\usepackage{epsfig}
\usepackage{graphicx}
\usepackage{amsmath}
\usepackage{amssymb}
\usepackage{tikz}
\usetikzlibrary{shadows,fit,positioning,backgrounds}
\usepackage[normalem]{ulem}
\usepackage{float}
\usepackage{subcaption}

\usepackage{multirow}

\usepackage[pagebackref=true,breaklinks=true,letterpaper=true,colorlinks,bookmarks=false]{hyperref}

\DeclareMathOperator*{\argmax}{argmax}

\cvprfinalcopy
\ifcvprfinal\fi
\begin{document}

\title{Top-down Visual Saliency Guided by Captions}

\author{Vasili Ramanishka\\
Boston University\\
{\tt\small vram@bu.edu}
\and
Abir Das\\
Boston University\\
{\tt\small dasabir@bu.edu}
\and
Jianming Zhang\\
Adobe Research\\
{\tt\small jianmzha@adobe.com}
\and
Kate Saenko\\
Boston University\\
{\tt\small saenko@bu.edu}
}

\maketitle
\thispagestyle{empty}

\begin{abstract}
Neural image/video captioning models can generate accurate descriptions, but their internal process of mapping regions to words is a black box and therefore difficult to explain. 
Top-down neural saliency methods can find important regions given a high-level semantic task such as object classification, but cannot use a natural language sentence as the top-down input for the task. 
In this paper, we propose Caption-Guided Visual Saliency to expose the region-to-word mapping in modern encoder-decoder networks and demonstrate that it is learned implicitly from caption training data, without any pixel-level annotations. Our approach can produce spatial or spatiotemporal heatmaps for both predicted captions, and for arbitrary query sentences.
It recovers saliency without the overhead of introducing explicit attention layers, and can be used to analyze a variety of existing model architectures and improve their design. 
Evaluation on large-scale video and image datasets demonstrates that our
approach achieves comparable captioning performance with existing methods while providing more accurate saliency heatmaps.
Our code is available at { \href{https://visionlearninggroup.github.io/caption-guided-saliency/}{visionlearninggroup.github.io/caption-guided-saliency/}}.
\end{abstract}

\section{Introduction}

\begin{figure}[t]
\begin{subfigure}{\linewidth}
\centering
\includegraphics[width=0.95\linewidth]{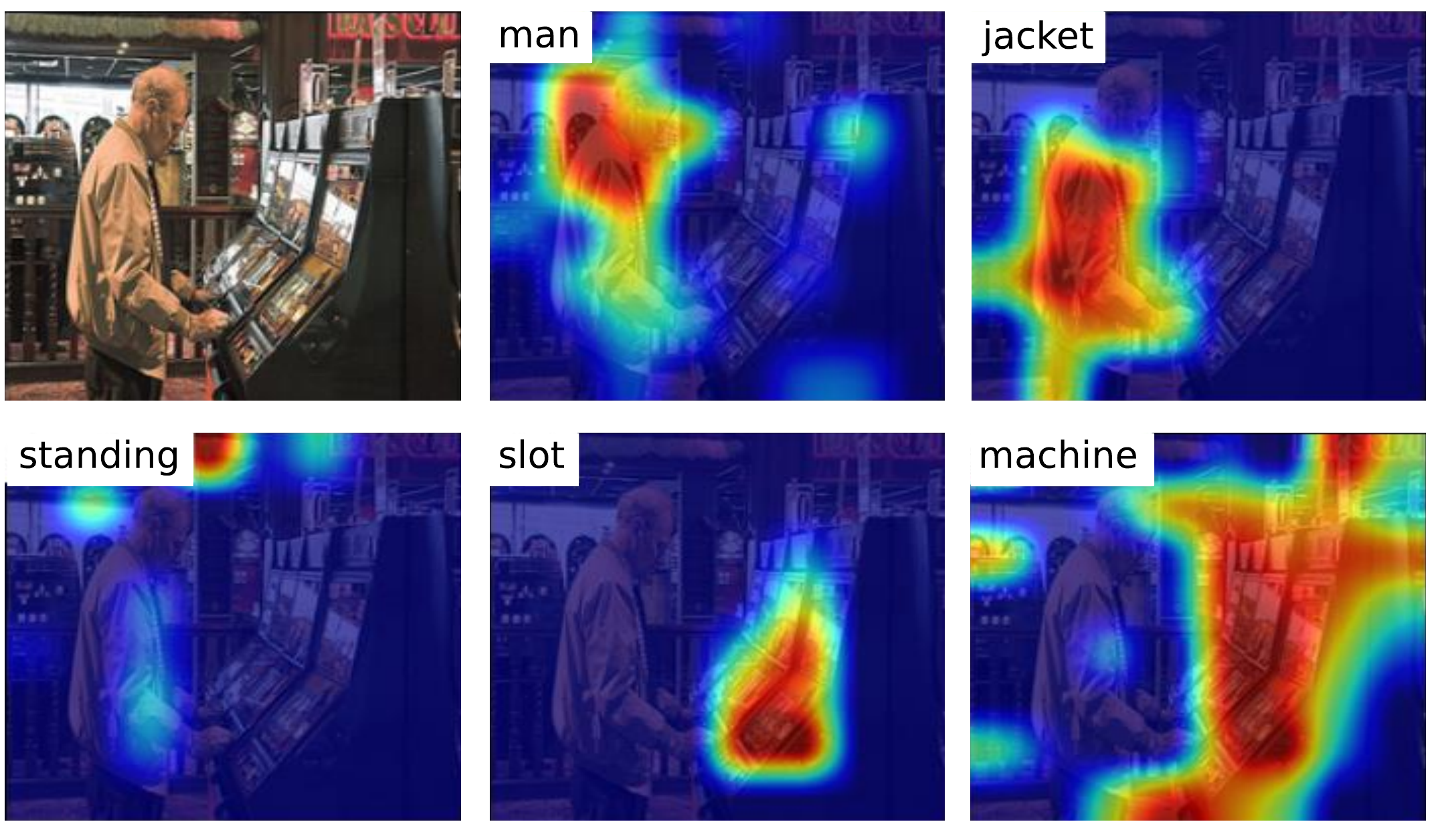}
\vspace{-0.1in}
\caption{\textbf{Input:} \textit{A man in a jacket is standing at the slot machine}}
\vspace{-0.1in}
\label{fig:slotmachine}
\end{subfigure}
\begin{subfigure}{\linewidth}
\centering
\includegraphics[width=0.96\linewidth]{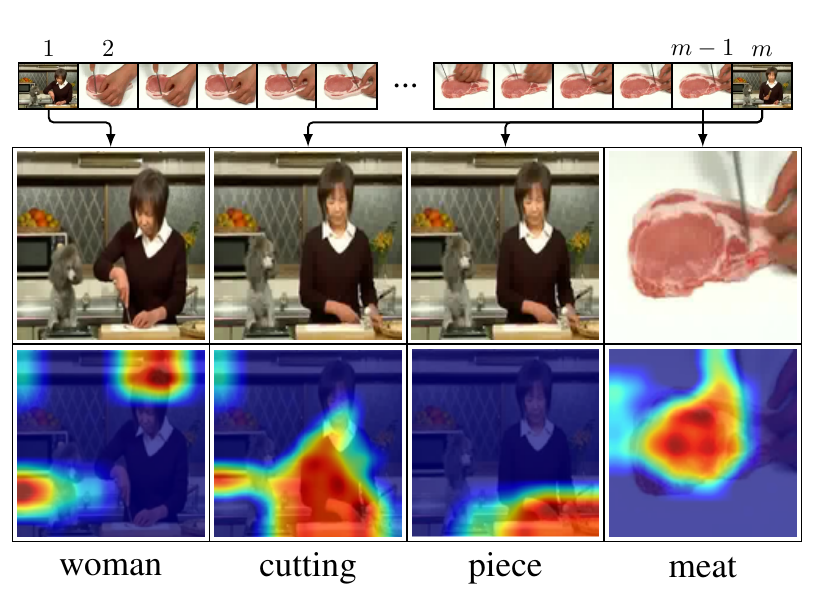}
\vspace{-0.1in}
\caption{\textbf{Input:} \textit{A woman is cutting a piece of meat}}
\vspace{-0.1in}
\label{fig:cuttingmeat}
\end{subfigure}
\caption{\small Top-down \textit{Caption-Guided Visual Saliency} approach that generates, for each word in a sentence, (a) spatial saliency in image and (b) spatiotemporal saliency in videos. For the video, we show temporally most important frames corresponding to the words at the bottom (arrows show positions of frames in the video) and spatial heatmaps indicating salient regions for these words.
}
\vspace{-0.3in}
\label{fig:MainFigure}
\end{figure}

Neural saliency methods have recently emerged as an effective mechanism for top-down task-driven visual search~\cite{Cao_2015_ICCV,ZhangECCV16}. They can efficiently extract saliency heatmaps given a high-level semantic input, \textit{e.g.}, highlighting regions corresponding to an object category, without any per-pixel supervision at training time. They can also explain the internal representations learned by CNNs~\cite{simonyan2013deep,zeiler2014visualizing}. However, suppose we wanted to  search a visual scene for salient elements described by a natural language sentence (Fig. 1(a)), or, given the description of an action, localize the most salient temporal and spatial regions corresponding to the subject, verb and other components (Fig. 1(b)). Classification-based saliency methods are insufficient for such language-driven tasks as they are limited to isolated object labels and cannot handle textual queries. 

Deep image and video captioning models~\cite{Chen2014LearningGeneration,Venugopalan2015,Vinyals2015ShowGenerator,Yao2015DescribingStructure} excel at learning representations that translate visual input into language potentially discovering a mapping between visual concepts and words. However, despite the good captioning performance, they can be very hard to understand and are often criticized for being highly non-transparent ``black boxes.'' 
They hardly provide any clear insight of the mapping learned internally
between the image and the produced words.
Consider for example, the video shown in Fig.~\ref{fig:MainFigure}(b). Which region in the model is used to predict words like ``woman'' or ``meat'' in the generated caption? Is the word ``woman'' generated because the model recognized the woman in the video, or merely because the language model predicts that ``A woman" is a likely way to start a sentence? Can the model learn to localize visual concepts corresponding to words while training only on weak annotations in the form of image or video-level captions? Can it localize words both in space and in time?

In this work, we address these questions by proposing a \textit{Caption-Guided Visual Saliency} method that leverages deep captioning models to generate top-down saliency for both images and videos. Our approach is based on an encoder-decoder captioning model, and can produce spatial or spatiotemporal heatmaps for either a given input caption or a caption predicted by our model (Fig.~\ref{fig:MainFigure}). In addition to facilitating visual search, this allows us to expose the inner workings of deep captioning models and provide much needed intuition of what these models are actually learning. This, in turn, can lead to improved model design in the future. Previous attempts at such model introspection have analyzed LSTMs trained on text generation~\cite{karpathy15}, or CNNs trained on image-level classification~\cite{ZhangECCV16,zhou2015cnnlocalization}.   
Recent ``soft'' attention models~\cite{Xu2015ShowAttention,Yao2015DescribingStructure} produce heatmaps by learning an explicit attention layer that weighs the visual inputs prior to generating the next word, but require modification of the network and do not scale well.
Thus, ours is the first attempt to analyze whether end-to-end visual captioning models can learn top-down saliency guided by linguistic descriptions without explicitly modeling saliency.

Our approach is inspired by the signal drop-out methods used to visualize convolutional activations in~\cite{zeiler2014visualizing,zhou2015cnnlocalization}, however we study LSTM based encoder-decoder models and design a novel approach based on information gain.
We estimate the saliency of each temporal frame and/or spatial region by computing the information gain it produces for generating the given word.
This is done by replacing the input image or video by a single region and observing the effect on the word in terms of its generation probability given the single region only.
We apply our approach to both still image and video description scenarios, adapting a popular encoder-decoder model for video captioning~\cite{2015SequenceText} as our base model.

Our experiments show that LSTM-based encoder-decoder networks can indeed learn the relationship between pixels and caption words. 
 To quantitatively evaluate how well the base model learns to localize words, we conduct experiments on the Flickr30kEntities image captioning dataset~\cite{flickr30kEntities}. 
We also use our approach to ``explain'' what the base video captioning model is learning on the publicly available large scale Microsoft Video-to-Text (MSR-VTT) video captioning dataset~\cite{Xu2016}. 
We compare our approach to explicit ``soft'' attention models~\cite{Xu2015ShowAttention,Yao2015DescribingStructure} and show that we can obtain similar text generation performance with less computational overhead, while also enabling more accurate localization of words.

\section{Related Work}

\noindent\textbf{Top-down neural saliency:}
Weak supervision in terms of class labels were used to compute the partial derivatives of CNN response with respect to input image regions to obtain class specific saliency map~\cite{simonyan2013deep}.
The authors in~\cite{zeiler2014visualizing} used deconvolution with max-pooling layers that projects class activations back to the input pixels.
While recent top-down saliency methods~\cite{Cao_2015_ICCV, mahendran2015understanding, ZhangECCV16, zhou2015cnnlocalization} recover pixel importance for a given class using isolated object labels, we extend the idea to linguistic sentences.

\noindent\textbf{Soft Attention:}
``Soft'' attention architectures, developed for machine translation~\cite{Bahdanau2014NeuralTranslate}, were recently extended to image captioning~\cite{Xu2015ShowAttention}.
Instead of treating all image regions equally, soft attention assigns different weights to different regions depending on the their content.
Similarly, in video captioning, an LSTM with a soft attention layer attends to specific temporal segments of a video while generating the description~\cite{Yao2015DescribingStructure}.
Compared to our top-down saliency model, one drawback of soft attention is that it requires an extra recurrent layer in addition to the LSTM decoder, requiring additional designing of this extra layer parameters.
The size of this layer scales proportionally to the number of items being weighted, \emph{i.e.}, the number of frames or spatial regions.
In contrast, our approach extracts the mapping between input pixels and output words from encoder-decoder models without requiring any explicit modeling of temporal or spatial attention and without modifying the network.
Our intuition is that LSTMs can potentially capture the inter-dependencies between the input and the output sequences through the use of memory cells and gating mechanisms.
Our framework visualizes both temporal and spatial attention without having to estimate additional weight parameters unlike explicit attention models, and can be used to analyse and provide explanations for a wide variety of encoder-decoder models.

\noindent\textbf{Captioning Models:}
Captioning models based on a combination of CNN and LSTM networks have shown impressive performance both for image and video captioning~\cite{Chen2014LearningGeneration, Venugopalan2015, Vinyals2015ShowGenerator, Yao2015DescribingStructure}. 
Dense captioning~\cite{Johnson16, Karpathy2015Deep} proposed to both localize and describe salient image regions.
Works on referring expression grounding~\cite{hu2016natural, mao2016generation, rohrbach2016grounding} localize input natural language phrases referring to objects or scene-parts in images.
These methods use ground truth bounding boxes and phrases to learn a mapping between regions and phrases.
We address the more difficult task of learning to relate regions to words and phrases without strong supervision of either, training only on images paired with their respective sentence captions.
We also handle spatiotemporal grounding for videos in the same framework.

\section{Background: Encoder-Decoder Model}\label{sec:background}

We start by briefly summarizing our base captioning model. We utilize the encoder-decoder video description framework~\cite{Venugopalan2015} which is based on sequence-to-sequence models proposed for neural translation~\cite{Cho2014LearningTranslation, Sutskever2014SequenceNetworks}. In Section~\ref{sec:Approach} we will describe how our approach applies the same base model to caption still images.

Consider an input sequence of $p$ video frames $\boldsymbol{x} = (x_{1}, \dots , x_p)$ and a target sequence of $n$ words $\boldsymbol{y} = (y_{1}, \dots, y_n)$.
The encoder first converts the video frames $\boldsymbol{x}$ into a sequence of $m$ high-level feature descriptors:

\begin{equation}
\label{eq:encodedSeq}
V = (\boldsymbol{v}_1, \dots, \boldsymbol{v}_m) = \phi(\boldsymbol{x})
\end{equation}
where typically $\phi()$ is a CNN pre-trained for image classification.
It then encodes the feature descriptors $V$ into a fixed-length vector $\boldsymbol{z} = E(\boldsymbol{v}_1, \dots, \boldsymbol{v}_m)$, where $E$ is some (potentially non-linear) function.
In the S2VT~\cite{2015SequenceText}, this is done by encoding $V$ into a sequence of hidden state vectors $\boldsymbol{h}^e_i$ using an LSTM, 
where the state evolution equation is:
\begin{equation}
\label{eq:encoderStateEvolution}
\vspace{-0.1in}
\boldsymbol{h}^e_{i} = f(\boldsymbol{v}_i, \boldsymbol{h}^e_{i-1}) \text{ for } i \in \{1, 2, \ldots, m\}
\end{equation}
and then taking $\boldsymbol{z}=\boldsymbol{h}^e_m$, the last LSTM state.
Another approach is to take the average of all $m$ feature descriptors~\cite{Venugopalan2015}, \textit{i.e.}, $\boldsymbol{z} = \frac{1}{m} \sum_{i=1}^{m} \boldsymbol{v}_i$.

The decoder converts the encoded vector $\boldsymbol{z}$ into output sequence of words $y_t$, $t \in \{ 1, \ldots, n\}$.
In particular, it sequentially generates conditional probability distribution for each element of the target sequence given encoded representation $\boldsymbol{z}$ and all the previously generated elements,
\vspace{-0.1in}
\begin{multline}
P(y_{t}|{y_{1}, \dots , y_{t-1}}, \boldsymbol{z}) = D(y_{t-1}, \boldsymbol{h}^d_{t}, \boldsymbol{z}), \\ \boldsymbol{h}^d_{t} = g(y_{t-1}, \boldsymbol{h}^d_{t-1}, \boldsymbol{z})
\label{eq:lstmDecoder}
\end{multline}
where $\boldsymbol{h}^d_{t}$ is the hidden state of the decoding LSTM and $g$ is again a nonlinear function.

\noindent\textbf{Soft Attention:} Instead of using the last encoder LSTM state or averaging $V$, the authors in~\cite{Yao2015DescribingStructure} suggest keeping the entire sequence $V$ and having the encoder compute a \textit{dynamic} weighted sum:
\vspace{-0.1in}
\begin{equation}
\hat{\boldsymbol{z}}_t = \sum_{i=1}^{m} \alpha_{ti}\boldsymbol{v}_i
\vspace{-0.1in}
\end{equation}
Thus, instead of feeding an averaged feature vector into the decoder LSTM, at every timestep a weighted sum of the vectors is fed.
The weights for every $\boldsymbol{v}_i$ are computed depending on previous decoder state $\boldsymbol{h}^d_{t-1}$ and encoded sequence $V=(\boldsymbol{v}_1, \dots, \boldsymbol{v}_m)$.
In video captioning, this allows for a search of related visual concepts in the whole video depending on the previously generated words.
As a result, one can think about attention in this model as a generalization of simple mean pooling across video frames.
Weights $\alpha_{ti}$ are obtained by normalizing $e_{ti}$, as follows,
\begin{equation}
\begin{gathered}
\label{eq:weighting_coeff}
\alpha_{ti} = \frac{exp(e_{ti})}{\sum_{k=1}^{m}exp(e_{tk})} \\
e_{ti} = \boldsymbol{w}^{T}tanh(\boldsymbol{W}_a\boldsymbol{h}_{t-1} + \boldsymbol{U}_{a}\boldsymbol{v}_i + \boldsymbol{b}_a)
\end{gathered}
\end{equation}
where $\boldsymbol{w}$, $\boldsymbol{W}_a$, $\boldsymbol{U}_a$ and $\boldsymbol{b}_a$ are attention parameters of the attention module. 

\begin{figure*}
\includegraphics[width=\textwidth]{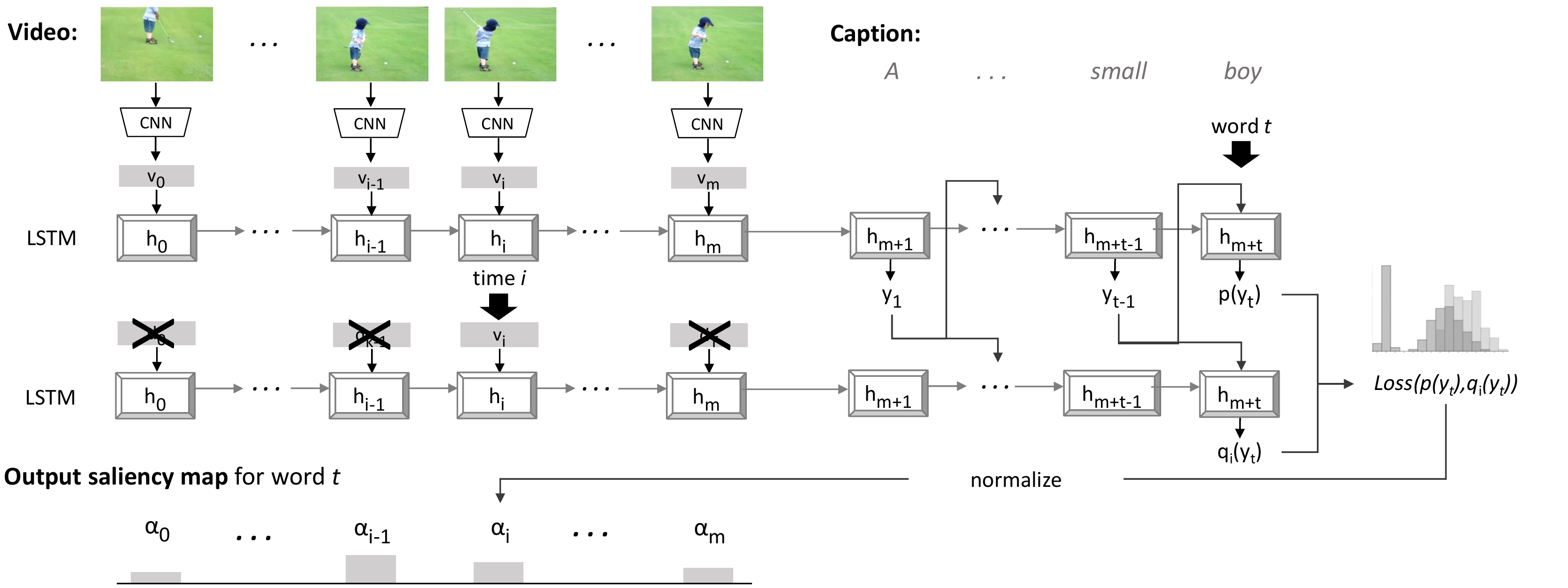}
\vspace{-0.1in}
\caption{\small Overview of our proposed top-down \textit{Caption-Guided Visual Saliency} approach for temporal saliency in video. We use an encoder-decoder model to produce temporal saliency values for each frame $i$ and each word $t$ in a given input sentence. The values are computed by removing all but the $i$th descriptor from the input sequence, doing a forward pass, and comparing to the original word probability distribution. A similar idea can be applied to spatial image saliency. See text for details.}
\vspace{-0.2in}
\label{fig:approach}
\end{figure*}

\section{Approach}
\label{sec:Approach}

We propose a top-down saliency approach called \textit{Caption-Guided Visual Saliency} which produces spatial and/or temporal saliency values (attention) for still images or videos based on captions. The saliency map can be generated for a caption predicted by the base model, or for an arbitrary input sentence. Our approach can be used to understand the base captioning model, \emph{i.e.} how well it is able to establish a correspondence between objects in the visual input and words in the sentence.
We use the encoder-decoder captioning model as our base model (equations~\ref{eq:encodedSeq}, \ref{eq:encoderStateEvolution}, \ref{eq:lstmDecoder}). 

For each word in the sentence, we propose to compute the saliency value of each item in the input sequence by measuring the decrease in the probability of predicting that word based on observing just that single item.
This approach is flexible, does not require augmenting the model with additional layers, and scales well with input size.
In contrast, in the soft attention model, the decoder selects relevant items from the input with the help of trainable attention weights.
This requires additional layers to predict the weights.
Furthermore, it can only perform either temporal or spatial mapping, but not both.
Our method estimates both a temporal and a spatial mapping between input and output using the base LSTM encoder-decoder model by recovering the \textit{implicit} attention from the model.
We describe the more general case of video in Section~\ref{sec:video} and then show how this model can be applied to still images in Section~\ref{sec:image}.

\subsection{Video Saliency}\label{sec:video}
In case of videos, we would like to compute the most salient spatiotemporal regions corresponding to words in the given sentence description of an event or activity.
Figure~\ref{fig:approach} shows an overview of the approach.
The intuition is that, although the encoder discards temporal and spatial positions of visual concept activations by encoding them into a fixed-length vector, this information can still be extracted from the model.
The encoded representation, containing activations of all visual concepts detected in the entire video, is passed on to the decoder LSTM at the start of the sentence generation process. The decoder then chooses parts of this state vector using LSTM output gates to predict the word at time $t$.
As each word is generated, the presence of visual concepts in the decoder LSTM state continually evolves, and the evolved state vector in turn interacts with the output gates to generate the next word.
As this interaction is complex and non-linear, we devise an indirect scheme to extract the evidence for the generation of each word.

Our approach measures the amount of information lost when a single localized visual input is used to approximate the whole input sequence.
The decoder predicts probability distributions $p(y_t)$ of words from the vocabulary at every step of the decoding process.
We assume that this probability distribution is our ``true'' distribution.
Then we measure how much information the descriptor of item $i$ carries for the word at timestep $t$.
To do this, we remove all descriptors from the encoding stage except for the $i$th descriptor.
After computing a forward pass through the encoder and decoder, this gives us a new probability distribution $q_i(y_t)$.
We then compute the loss of information as the KL-divergence between the two distributions, 
\begin{align}
\label{eq:loss}
p(y_t) &= P(y_t | y_{1:t-1}, v_{1:m}) \nonumber\\
q_i(y_t) &= P(y_t | y_{1:t-1}, v_{i}) \\
Loss(t,i) &= D_{KL}(p(y_t) \lVert q_i(y_t)) \nonumber
\end{align}

With the above formulation we can easily derive top-down saliency for word $w$ predicted at time $t$. We assume that the query sentence $S$ has ``one-hot'' ``true'' distributions on every timestep. With this assumption Eq.~\ref{eq:loss} reduces to:
\vspace{-0.1in}
\begin{align}
\label{eq:loss2}
Loss(t,i,w) &= \sum_{k \in W} p(y_t = k)\log\frac{p(y_t = k)}{q_i(y_t = k)} \nonumber \\
         &= \log\frac{1}{q_{i}(y_t = w)} 
\end{align}
This process is not limited to produced word sequence only but can be used with any arbitrary query for a given video.

As the approximate receptive field of each descriptor can be estimated\footnote{for our video descriptors, the receptive field is a single frame},
we can define a saliency map for each word in the sentence by mapping $Loss(t,i,w)$ to the center of the receptive field and and upsampling the resulting heatmap. 
It follows from Eq.~\ref{eq:loss2} that $Loss(t, i, w) \in [0; +\infty)$, where values which are closer to zero correspond to higher saliency. 
To obtain a saliency value $e_{ti}$, we negate the loss and linearly scale the resulting values to the $[0, 1]$ interval, 
\begin{align}
\label{eq:eti}
e_{ti} = scale(-Loss(t,i,w))
\end{align}

It is important to discriminate between the values meant by Eq.~\ref{eq:loss} and.~\ref{eq:loss2}. The former can be used to evaluate the representativeness of individual descriptors compared to the full input sequence, while the latter induces top-down saliency maps for individual words at each time step. 
Finally, the saliency value for a group of words from the target sentence (e.g. a noun phrase ``a small boy'') is defined as sum of the corresponding saliency values for every word in the subsequence:
\vspace{-0.1in}
\begin{equation}
\label{eq:loss3}
Loss(\{t_1, ..., t_{q}\},i) = \sum_{j=1}^{q} Loss(t_j, i).
\end{equation}

Next we describe how this approach is applied to generate both temporal and spatial saliency in videos.

\textbf{Temporal attention:}
For an input frame sequence $V=(\boldsymbol{v}_1, \dots, \boldsymbol{v}_m)$, the deterministic algorithm of sentence generation is given by the following recurrent relation:
\begin{equation}
w = \argmax_{y_t \in W}{p(y_t|y_{0:t-1}, \boldsymbol{v}_{1:m}})
\end{equation}
where $y_0$ and $y_n$ are special ``begin of sequence'' and ``end of sequence'' tokens respectively.
Given the word predicted at time $t$ of the sentence, the relative saliency of the input frame $\boldsymbol{v}_i$  can be computed as $e_{ti}$ (Eq.~\ref{eq:eti}).
In other words, we estimate the drop in probability of every word in the output sequence resulting from encoding only that input frame. Further, we normalize $\textbf{e}_t=(e_{t1}, \dots, e_{tm})$ to obtain stochastic vectors as in Eq.~\ref{eq:weighting_coeff} and interpret the resulting vectors $\boldsymbol{\alpha}_{t}=(\alpha_{t1}, \dots, \alpha_{tm})$ as saliency over the input sequence  $V =(\boldsymbol{v}_1, \dots, \boldsymbol{v}_m)$ for every word $y_{t}$ of the output sequence.
This also induces a direct mapping between predicted words and the most salient frames for these words. 

\textbf{Spatial attention:}
We can also estimate the attention on different frame patches as related to a particular word $y_t$ of a sentence.
Although spatial pooling in the CNN discards the spatial location of detected visual concepts, the different gates of the LSTM enable it to focus on certain concepts depending on the LSTM hidden state.
Let $f_k(a,b)$ be the activation of unit $k$ (corresponding to some visual concept) at spatial location $(a,b)$ in the last convolutional layer of the encoder~\cite{zhou2015cnnlocalization}.
The CNN performs spatial average pooling to get a feature vector $\boldsymbol{v}_i$ for the $i^{th}$ frame whose $k^{th}$ element is $v_{ik} = \sum_{a,b}f_k(a,b)$.
After that, the encoder embeds the descriptor into LSTM cell state according to the LSTM update rule. This process involves the LSTM input gate:
\begin{equation}
\label{eq:inputGate}
\rho_i = \sigma(W_{v\rho}\boldsymbol{v}_i + W_{h\rho}h_{i-1} + b_{\rho})
\end{equation}
where the LSTM selects activations $\boldsymbol{v}_{ik}$ by weighting them depending on the previous LSTM hidden state and $\boldsymbol{v}_i$ itself ($W_{v\rho}$, $W_{h\rho}$ and $b_{\rho}$ are trainable parameters).
Note that,
\begin{equation} 
\label{eq:someMath}
\hspace{-3mm}
\small
W_{v\rho}\boldsymbol{v}_{i}\!\!=\!\!\sum_k{\!\!\boldsymbol{w}_{k} \boldsymbol{v}_{ik}} \! = \!\!\! \sum_k{\!\!\boldsymbol{w}_k \!\! \sum_{a,b} \!\! f_k(a,b)} \! = \!\! \sum_{a,b}{\! \sum_k{ \! \boldsymbol{w}_k f_k(a,b)}}
\vspace{-0.1in}
\end{equation}
where $\boldsymbol{w}_k$ denotes the $k^{th}$ column of matrix $W_{v\rho}$. Since each unit activation $f_k(a,b)$ represents a certain visual concept~\cite{zeiler2014visualizing}, we see that the input gate learns to select input elements based on relevant concepts detected in the frame, regardless of their location.
The explicit spatial location information of these concepts is lost after the spatial average pooling in the last convolutional layer,
however, we can recover it from the actual activations $f_k(a,b)$.
This is achieved by computing the information loss for different spatial regions in a frame in a similar way as was done for temporal attention extraction.
The relative importance of region $(a,b)$ in frame $\boldsymbol{v}_i$ for word $w$ predicted at time $t$ can be estimated as: 
\begin{align}
e_{ti}^{(a,b)} = - Loss(t,i,w), \nonumber\\
\text{where}~~ p(y_t)=P(y_t|y_{0:t-1}, \boldsymbol{v}_{1:m}),  \\  q_i(y_t)=P(y_t|y_{0:t-1},\boldsymbol{v}_{i}^{(a,b)} ),\nonumber
\label{eq:spatialAttention}
\end{align}
and where $\boldsymbol{v}_{ik}^{(a,b)}= f_k(a,b)$.
Assuming the number of spatial locations in a frame to be $r$, the prediction process (i.e. forward pass) is run $m$ times to obtain temporal saliency maps and $r \times m$ times to obtain the spatial maps for the given video/sentence pair. This, in turn, involves $n+1$ LSTM steps, so the total complexity is
\begin{equation}
O(
    (
    \underbrace{r \times m + m}_{\text{spatial and temporal }} 
    ) 
 \times
    (
    \underbrace{n + 1}_{\text{LSTM steps}}
    )
 )
\end{equation}

Since all $Loss(t, i, w)$ computations are performed independently, we can create a batch of size $r \times m + m$ and calculate all the saliency values efficiently in one pass. 

\begin{figure*}[!t]
\centering
\begin{subfigure}{1.0\textwidth}
\centering
\includegraphics[width=0.95\linewidth]{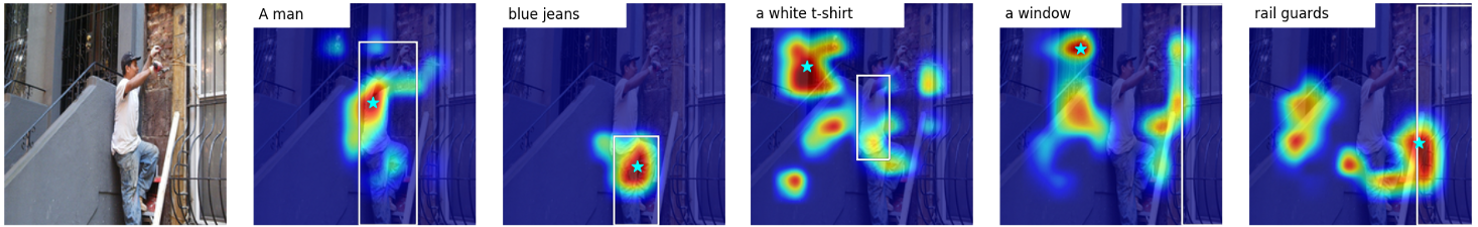}
\caption{\small A man in blue jeans and a white t-shirt is working on a window with rail guards.}
\label{fig:01}
\end{subfigure}
\begin{subfigure}{1.0\textwidth}
\centering
\includegraphics[width=0.95\linewidth]{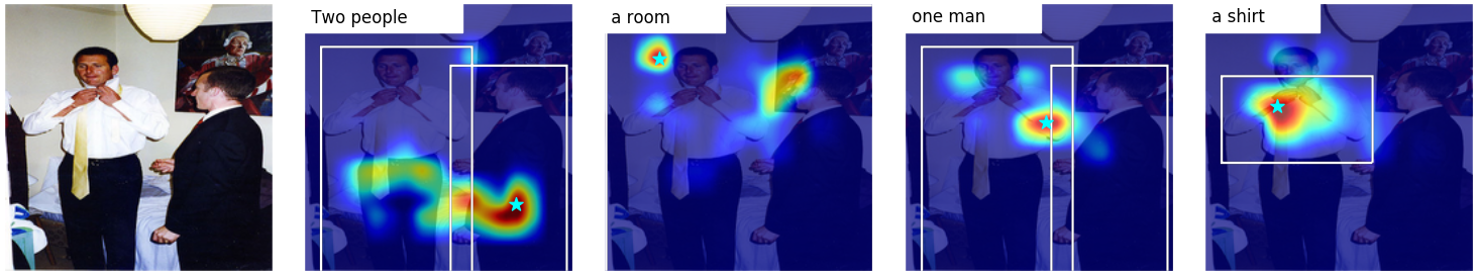}
\caption{\small Two people are in a room , one man is putting on a shirt and tie.}
\label{fig:02}
\end{subfigure}
\begin{subfigure}{1.0\textwidth}
\centering
\includegraphics[width=0.95\linewidth,height=0.1\paperheight]{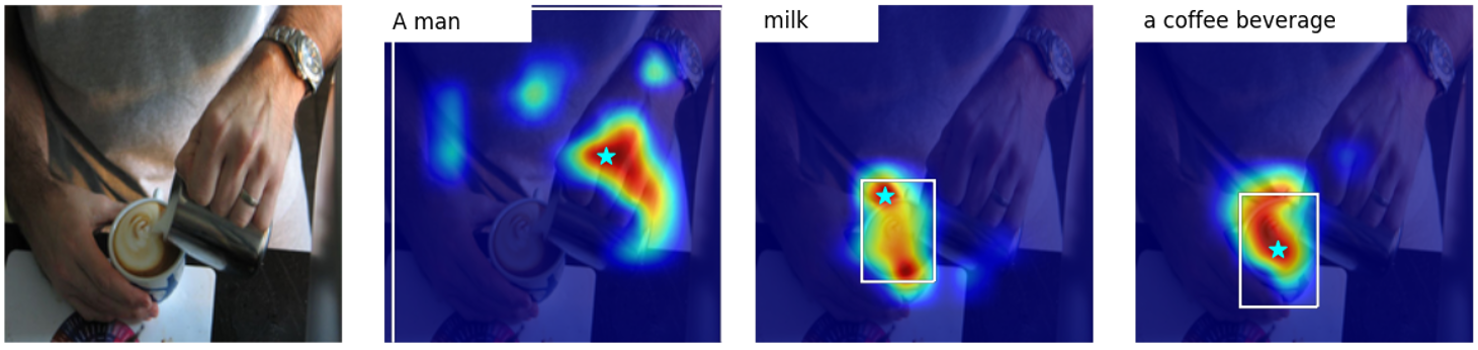}
\caption{\small A man is adding steamed milk to a coffee beverage.}
\label{fig:03}
\end{subfigure}
\begin{subfigure}{1.0\textwidth}
\centering
\includegraphics[width=0.95\linewidth,height=0.1\paperheight]{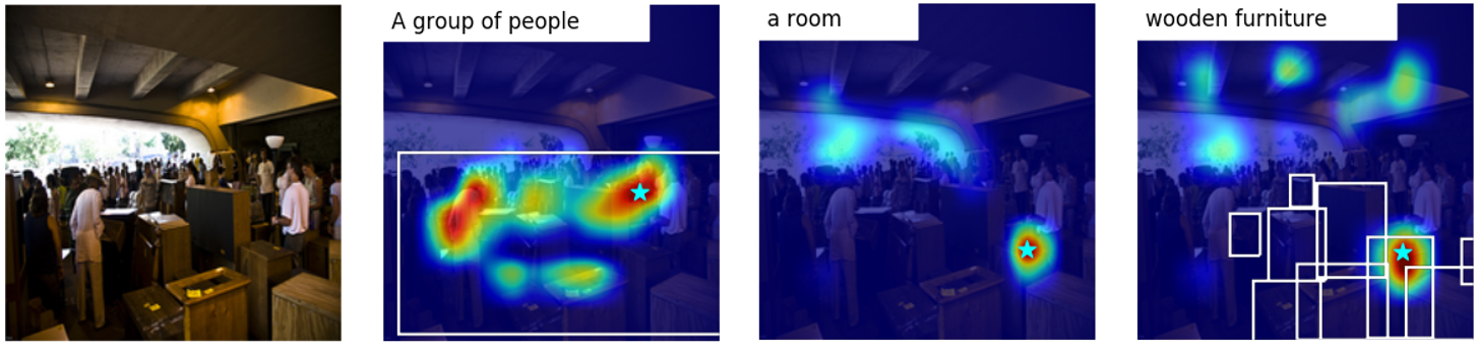}
\caption{\small A group of people are standing in a room filled with wooden furniture.}
\label{fig:04}
\end{subfigure}
\vspace{-0.1in}
\caption
{\small Saliency maps (red to blue denotes high to low value) in Flickr30kentities generated for an arbitrary query sentence (shown below). Each row shows saliency map for different noun-phrases (shown at top-left corner) extracted from the query. Maximum saliency point is marked with asterisk and ground truth boxes are shown in white.
}
\vspace{-0.1in}
\label{fig:visualization_query}
\end{figure*}

\begin{figure*}[!t]
\centering
\includegraphics[width=0.95\linewidth]{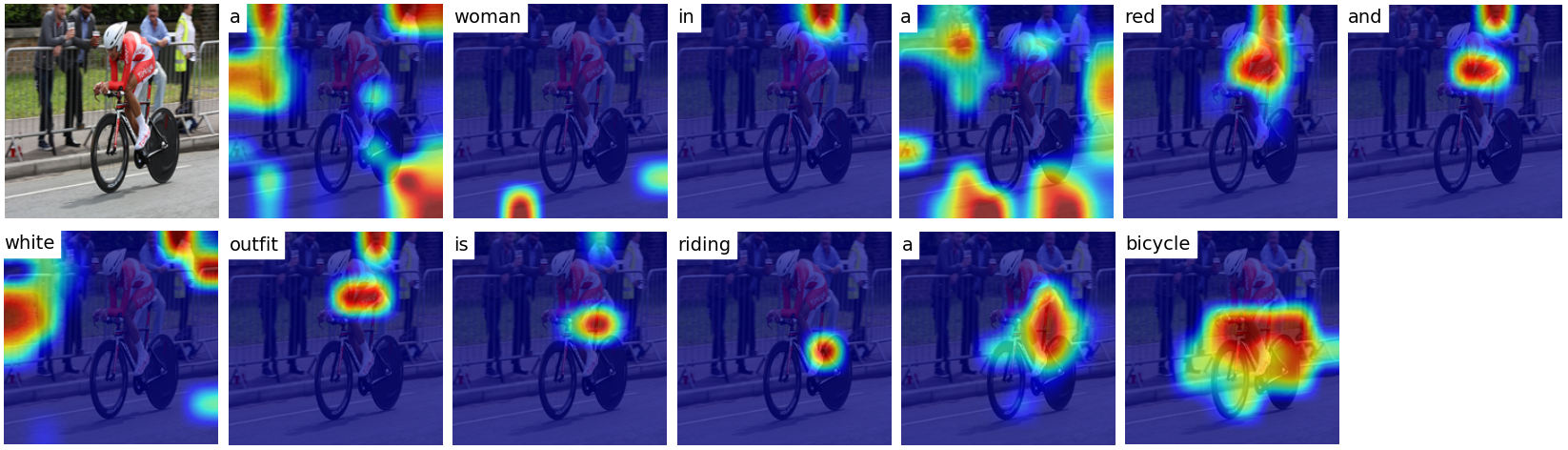}
\text{\small A woman in a red and white outfit is riding a bicycle.}
\vspace{-0.1in}
\caption
{\small Saliency maps generated for a caption (shown below the image) predicted by the model. 
}
\vspace{-0.1in}
\label{fig:visualization_predicted}
\end{figure*}

\subsection{Image Saliency}\label{sec:image}
With minimal changes the above model can be applied to generate saliency for images.
We accomplish this by re-arranging the grid of descriptors produced by the last convolutional layer of the CNN into a ``temporal'' sequence $V = (\boldsymbol{v}_1, \dots, \boldsymbol{v}_m)$ by scanning the image in a sequential manner (row by row), starting from the upper left corner and ending at the bottom right corner.
Our model uses the encoder LSTM to scan the image locations and encode the collected visual information into hidden states and then decodes those states into the word sequence.
Generating a spatial saliency map can now be achieved by the same process as described for temporal saliency in the previous section.


\begin{table*}[t]
\centering
\caption{\small Evaluation of the proposed method on localizing all noun phrases from the ground truth captions in the Flickr30kEntities dataset using the pointing game protocol from~\cite{ZhangECCV16}. ``Baseline random'' samples the point of maximum saliency uniformly from the whole image and ``Baseline center'' corresponds to always pointing to the center.}
\label{tab:quantitativeResultsPointingGame}
\vspace{-0.1in}
\begin{tabular}{@{}llllllllll@{}}

                                  & bodyparts & animals        & people & instruments & vehicles & scene          & other          & clothing & Avg per NP     \\\hline
Baseline random                   & 0.100     & 0.240          & 0.318  & 0.179       & 0.275    & 0.524          & 0.246          & 0.151    & 0.268          \\
Baseline center                   & \textbf{0.201}     & 0.599          & \textbf{0.647}  & \textbf{0.496}       & 0.644    & 0.652          & 0.384          & \textbf{0.397}    & 0.492 \\
Our Model                         & 0.194     & \textbf{0.690} & 0.601  & 0.458       & \textbf{0.645}    & \textbf{0.667} & \textbf{0.427} & 0.360    & \textbf{0.501} \\\hline
\end{tabular}\label{tab:resultsPointingGame}
\end{table*}

\begin{table*}[t]
\centering
\caption{\small Evaluation of the proposed method on Flickr30kEntities using the \emph{attention correctness metric} and evaluation protocol from~\cite{liu2016attentioncorrectness} (including the frame cropping procedure). Soft attention performance is taken from~\cite{liu2016attentioncorrectness} as reported there.  \textit{Baseline*} shows our re-evaluation of the uniform attention baseline.}
\vspace{-0.1in}
\label{quantitativeResults}
\begin{tabular}{@{}lccccccccc@{}}
                                  & bodyparts & animals        & people & instruments & vehicles & scene          & other          & clothing & Avg per NP     \\\hline
Baseline~\cite{liu2016attentioncorrectness}                         & -         & -              & -      & -           & -        & -              & -              & -        &  0.321    \\
Soft attention~\cite{Xu2015ShowAttention}                    		& -         & -              & -      & -           & -        & -              & -              & -        & 0.387         \\
Soft attention             & \multirow{2}{*}{-}         & \multirow{2}{*}{-}              & \multirow{2}{*}{-}      & \multirow{2}{*}{-}          & \multirow{2}{*}{-}       & \multirow{2}{*}{-}             &\multirow{2}{*}{-}              & \multirow{2}{*}{-}       & \multirow{2}{*}{0.433}         \\
supervised~\cite{liu2016attentioncorrectness}             &          &              &      &           &        &              &              &     &          \\
\hline

Baseline*       & 0.100        & 0.371          & 0.410        & 0.278        & 0.350        & 0.470          & 0.236          & 0.197        & 0.325 \\
Our model                   & \textbf{0.155}        & \textbf{0.657}          & \textbf{0.570}        & \textbf{0.502}        & \textbf{0.615}        & \textbf{0.582}          & \textbf{0.348}          & \textbf{0.345}        & \textbf{0.473}          \\ 
\hline
\end{tabular}\label{tab:quantitativeResultsAttentionCorrectness}
\end{table*}

\section{Experiments}
\label{sec:Experiments}
This section shows examples of caption-driven saliency recovered by our method for the base S2VT model from videos and still images. We evaluate the quality of the recovered heatmaps on an image dataset annotated with ground truth object bounding boxes. 
We also evaluate the caption generation performance on both images and videos and compare it with the soft attention approach. 

\begin{figure}[!t]
\centering
\includegraphics[width=\linewidth]{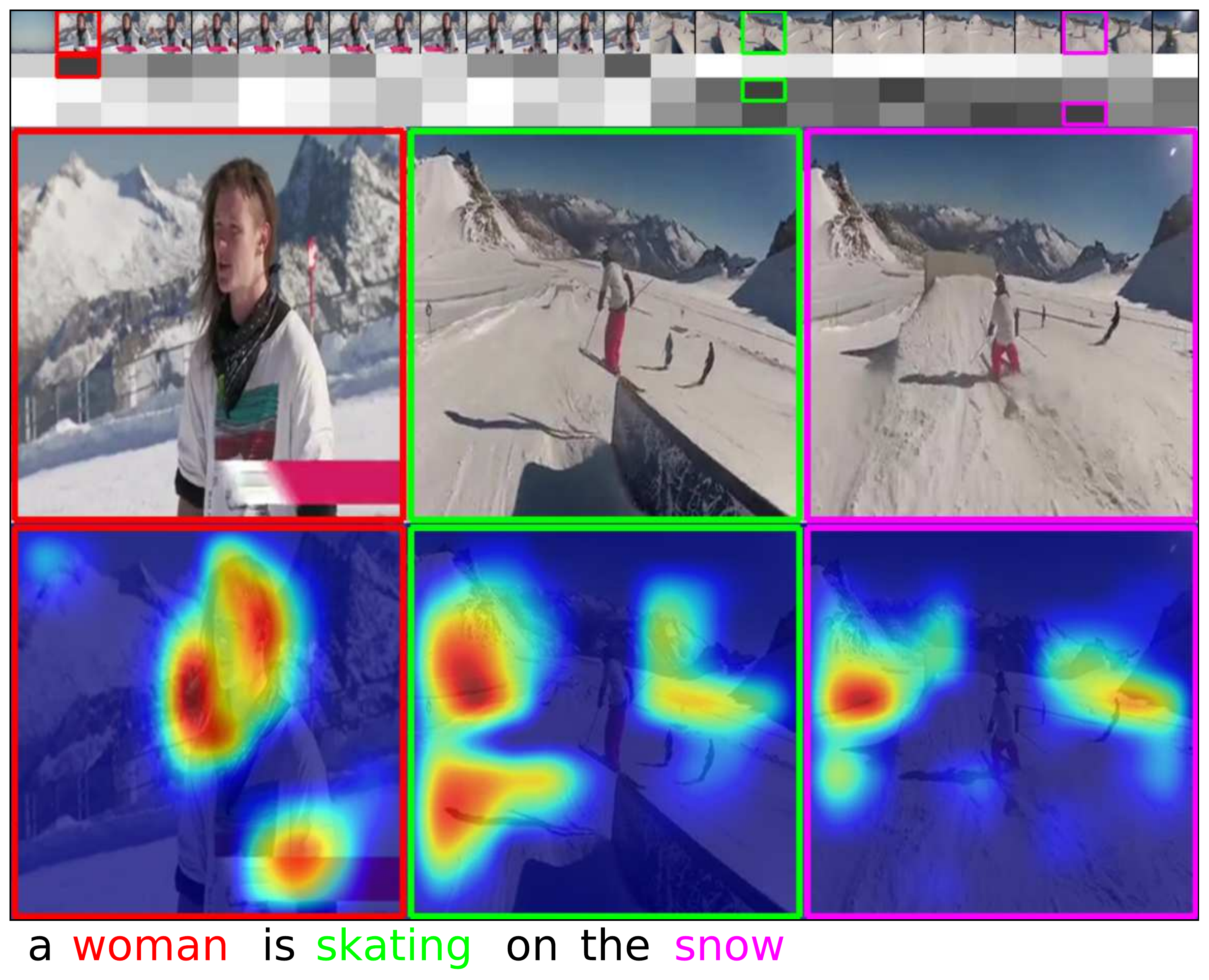}
\includegraphics[width=\linewidth]{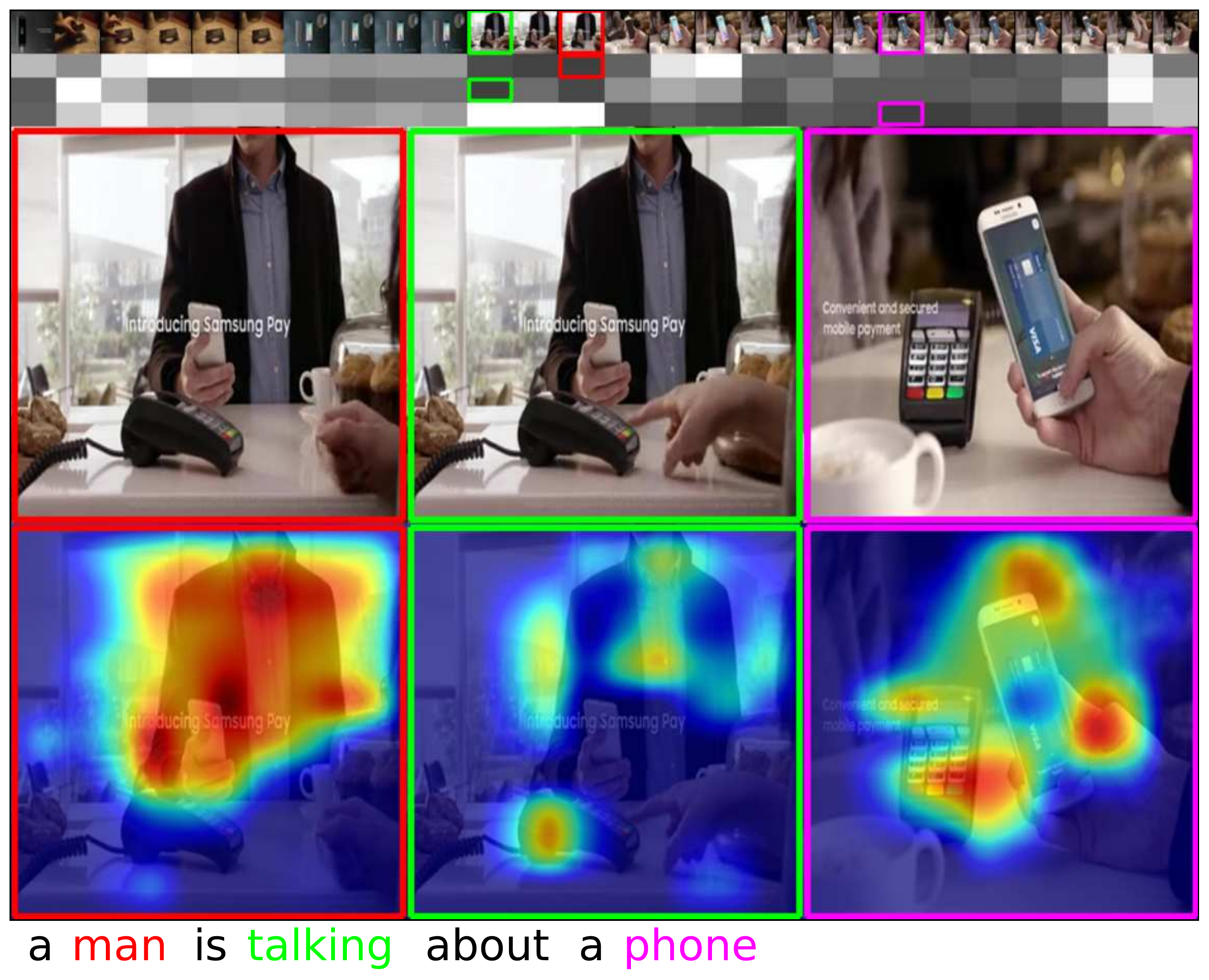}
\vspace{-0.3in}
\caption{\small Spatial and temporal saliency maps in videos. For each word, darker grey indicates higher relative saliency of the frame. For better visualization, saliency values are not normalized but linearly mapped to the range [0, 1].
Most relevant frames for each word are shown at the bottom, highlighted with the same color.}
\vspace{-0.1in}
\label{fig:temporalAttention}
\end{figure}

\noindent \textbf{Datasets}
We train and evaluate our model on two video description datasets, namely the Microsoft Video Description dataset (MSVD)~\cite{chen2011collecting} and the Microsoft Research Video to Text (MSR-VTT)~\cite{Xu2016} dataset.
Both datasets have ``in the wild" Youtube videos and natural language descriptions.
MSVD contains 1970 clips of average length 10.2s with 80,827 natural language descriptions.
MSR-VTT provides 41.2 hours of web videos as 10,000 clips of approx. 14.8s each and 200K natural language descriptions. 
In addition, we evaluated on one of the largest image captioning datasets, Flickr30kEntities~\cite{flickr30kEntities} which is an extension of the original Flick30k~\cite{TACL229} dataset with manual bounding box annotations for all noun phrases in all 158k image captions.

\noindent \textbf{Model details}
We implemented our model in TensorFlow~\cite{tensorflow2015-whitepaper} using InceptionV3~\cite{Szegedy2016Rethinking} pretrained on ImageNet~\cite{imagenet_cvpr09} as CNN feature extractor.
We use ${v_1, \dots, v_{26}}, v_i \in \mathbb{R}^{2048}$ for the video representation. $v_i$ were extracted from the average pooling layer~\textit{pool\_3} for $26$ evenly spaced frames. For images we use feature outputs from the last convolutional layer \textit{mixed\_10} as the input sequence to the encoder. 
Thus, for video and image captioning the input sequences have length $m=26$ and $m=64$ respectively. The order of spatial descriptors for image captioning is described in Section~\ref{sec:image}.
All images and video frames were scaled to 299x299. 
Note that the CNN was trained on ImageNet and was not finetuned during the training of the captioning model. 
A fully-connected layer reduced the dimensionality of the input descriptors from 2048 to 1300 before feeding them into the LSTM.
The model was trained using the Adam optimizer with initial learning rate 0.0005. Dimensionality of the word embedding layer was set to 300.

\noindent \textbf{Evaluation of captioning performance}
Quantitative evaluation of the captioning performance was done using the METEOR~\cite{Banerjee2005} metric.
Table~\ref{table:params} (higher numbers are better) shows the results and demonstrates that despite not using explicit attention layers, our model performs comparably to the soft attention method.
The best model in terms of the METEOR metric on the validation split of Flickr30k was selected for the evaluation of saliency as presented below. 

\noindent\textbf{Quantitative evaluation of saliency}
Given a pretrained model for image captioning, we test our method quantitatively using the~\textit{pointing game} strategy~\cite{ZhangECCV16} and \textit{attention correctness metric}~\cite{liu2016attentioncorrectness}. To generate saliency maps, we feed ground truth captions from the test split of Flickr30k into our model.
In pointing game evaluation, we obtain the maximum saliency point inside the image for each annotated noun phrase in each GT caption of Flickr30kEntities.
We then test whether this point lies inside the bounding box or not.
Accuracy is computed as $Acc=\frac{\#Hits}{\#Hits + \#Misses}$.
To get a saliency map for noun phrases which are comprised of multiple tokens from the sentence, we sum loss values before their normalizing them to the $[0, 1]$. 

Table~\ref{tab:quantitativeResultsPointingGame} shows the mean accuracy over all noun phrases (NPs) along with accuracies corresponding to categories (in different columns) from Flickr30kEntities.
We compare to ``Baseline random'', where the maximum saliency point is sampled uniformly from the whole image and to a much stronger baseline denoted as ``Baseline center''.
This baseline is designed to mimic the center bias present in consumer photos and assumes that the maximum saliency point is always at the center of the image.
Compared to the random baseline, the accuracy of the proposed method is better on average (last column) as well as for all the individual categories (rest of the columns).
While the average accuracy compared to the much stronger center baseline is only slightly better, the accuracy gain for some of the categories is significant.
One possible reason may be that the objects in these categories, \textit{e.g.}, `animals' or `other' objects, tend to be away from the central region of an image, while people tend to be in the center of the photo.

Table~\ref{tab:quantitativeResultsAttentionCorrectness} provides a direct comparison of our method to the soft attention model~\cite{Xu2015ShowAttention} in terms of the \textit{attention correctness metric} proposed in~\cite{liu2016attentioncorrectness}.
This metric measures average value for integral of attention function over bounding boxes.
We directly report the results from~\cite{liu2016attentioncorrectness} for their implementation of uniform baseline, soft-attention model and its improved version where a captioning model was trained to focus on relevant objects in supervised manner.
Our method outperforms all three of them.

We also provide the category specific values as we obtained from our own implementation of the uniform baseline (called ``Baseline*'').
``Baseline random'' in Table~\ref{tab:quantitativeResultsPointingGame} should roughly correspond to ``Baseline'' and ``Baseline*" in Table~\ref{tab:quantitativeResultsAttentionCorrectness}.
Evidently, the exact values will be different as the evaluation protocols in the two tables are different.
To compare the results fairly, we followed the same protocol as~\cite{liu2016attentioncorrectness} where the authors performed a central crop of both test and training images.
Human-captured images or videos tend to put the objects of interest in the central region.
Thus, any cropping operation which enhances this ``central tendency'' will, inherently, give a better measure of attention.
This frame cropping strategy is another source of discrepancy in the baseline values in Table~\ref{tab:quantitativeResultsPointingGame} and~\ref{tab:quantitativeResultsAttentionCorrectness}.

\begin{table}
\begin{center}
\small
\caption{\small Comparison of the captioning performance of our model and soft-attention on two video (MSVD, MSR-VTT) and one image (Flickr30k) datasets. Higher numbers are better.}
\vspace{-0.1in}
\label{table:params}
\begin{tabular}{llc}
                            Model & \ Dataset &METEOR~\cite{denkowski2014meteor}  \\ \hline
Soft-Attn~\cite{Yao2015DescribingStructure} 			& MSVD  & 30.0\\
Our Model 							  					& MSVD   & 31.0\\ 
 \hline
Soft-Attn~\cite{msr-vttSuppl} & MSR-VTT  & 25.4\\
Our Model & MSR-VTT  & 25.9\\
 \hline
Soft-Attn~\cite{Xu2015ShowAttention} & Flickr30k  & 18.5\\
Our Model & Flickr30k  & 18.3\\
\vspace{-0.3in}
\end{tabular}
\end{center}
\end{table}

\noindent\textbf{Saliency visualizations in images} 
Figures~\ref{fig:visualization_query} and \ref{fig:visualization_predicted} show example saliency maps on images from Flickr30kEntities for arbitrary query sentences and model-predicted captions, respectively.
The arbitrary query comes from the ground truth descriptions.
For each nounphrase, the saliency map is generated by summing the responses for each token in the phrase and then renormalizing them.
The map is color coded where red shows the highest saliency while blue is the lowest.
The maximum saliency point is marked with an asterisk, while the ground truth boxes for the noun-phrases are shown in white.
It can be seen that our model almost always localizes humans correctly.
For some other objects the model makes a few intuitive mistakes.
For example, in Fig.~\ref{fig:01}, though the saliency for ``window'' is not pointing to the groundtruth window, it focuses its highest attention (asterisk) on the gate which looks very similar to a window.
In Fig.~\ref{fig:visualization_predicted}, the saliency map the predicted caption fof an image is shown.
Some non-informative words (\textit{e.g.}, ``a'', ``is'' \textit{etc.}) may appear to have concentrated saliency, however, this is merely a result of normalization.
One surprising observation is that the model predicts `a woman in a red and white outfit', however only the `red' spatial attention is on the cyclist, while the `white' attention is on other parts of the scene.

\noindent\textbf{Saliency visualizations in videos}
Fig.~\ref{fig:temporalAttention} shows examples of spatial and temporal saliency maps for videos from MSR-VTT dataset with model-predicted sentences.
Most discriminative frames for each word are outlined in the same color as the word.
Darker gray indicates higher magnitude of temporal saliency for the word.
We omit visualization for uninformative words like articles, helper verbs and prepositions.
An interesting observation about the top video is that the most salient visual inputs for ``skating'' are regions with snow, with little attention on the skier, which could explain the mistake.

Additional results and source code are available at \href{https://visionlearninggroup.github.io/caption-guided-saliency/}{visionlearninggroup.github.io/caption-guided-saliency/}.

\section{Conclusion}
We proposed a top-down saliency approach guided by captions and demonstrated that it can be used to understand the complex decision processes in image and video captioning without making modifications such as adding explicit attention layers.
Our approach maintains good captioning performance while providing more accurate heatmaps than existing methods.
The model is general and can be used to understand a wide variety of encoder-decoder architectures. 

\section{Acknowledgements}
This research was supported in part by NSF IIS-1212928, DARPA, Adobe Research and a Google Faculty grant.
We thank Subhashini Venugopalan for providing an implementation of S2VT~\cite{2015SequenceText} and Stan Sclaroff for many useful discussions. 

{\small
\bibliography{references}
\bibliographystyle{abbrv}
}

\end{document}